\documentclass[a4paper,12pt]{article}
\usepackage{arxiv}

\usepackage{graphicx}
\usepackage[numbers,square]{natbib}
\usepackage{amsmath}
\usepackage{color}
\usepackage{amssymb}
\usepackage{float}
\usepackage{array}
\usepackage{wrapfig}

\newcolumntype{C}[1]{ >{\centering\arraybackslash} m{#1} }

\graphicspath{ {images/} }

\title{Reducing catastrophic forgetting when evolving neural networks}
\author{
	Joseph Early \\
	Electronics and Computer Science\\
	University of Southampton \\
	\texttt{je5g15@soton.ac.uk} \\
}

\begin{document}
\maketitle

\begin{abstract}
A key stepping stone in the development of an artificial general intelligence (a machine that can perform any task), is the production of agents that can perform multiple tasks at once instead of just one. Unfortunately, canonical methods are very prone to catastrophic forgetting (CF) - the act of overwriting previous knowledge about a task when learning a new task. Recent efforts have developed techniques for overcoming CF in learning systems, but no attempt has been made to apply these new techniques to evolutionary systems. This research presents a novel technique, weight protection, for reducing CF in evolutionary systems by adapting a method from learning systems. It is used in conjunction with other evolutionary approaches for overcoming CF and is shown to be effective at alleviating CF when applied to a suite of reinforcement learning tasks. It is speculated that this work could indicate the potential for a wider application of existing learning-based approaches to evolutionary systems and that evolutionary techniques may be competitive with or better than learning systems when it comes to reducing CF.
\end{abstract}


\section{Introduction}

One of the main divides between machine intelligence and the only known example of general intelligence, human intelligence, is a lack of generality. The human brain can adapt to any task with relative ease, and learns over a much longer timescale than its artificial counterpart. Indeed, the human brain is always learning or developing in some way, yet it is remarkably resilient against forgetting what it has previously learnt. The ease at which human intelligence can swap tasks and remember former skills is something that is thus far beyond the grasp of machine intelligence.

A key stepping stone in the development of an artificial general intelligence (a machine that can perform any task), is the production of agents that can perform multiple tasks instead of just one. This is the notion of continual learning (CL) - learning task A then learning task B and so on  \cite{ring1994continual}. Unfortunately, simply training an agent on a new task after it has already learnt a different task will likely result in a serious decrease in performance on the original task; this is known as catastrophic forgetting (CF) \cite{French1999}.

For an agent to be useful in the real world, or sufficiently rich world like simulations, it must be able to interpret signals from the world around it and interact with it in some way. This process of agent interaction within some environment is known as reinforcement learning (RL). In an RL scenario, an agent transitions between states by taking actions; the possible states and actions in those states make up the environment that the agent is in. The goal of the agent is to maximise some cumulative reward function over a number of discrete time steps \cite{sutton1998reinforcement}. 

This research covers two approaches to solving RL problems: learning based and evolutionary based. In learning approaches, a single agent is developed, where the agent has very poor performance initially but improves over time by learning from its mistakes in a trial and error fashion. In evolutionary approaches, a population of agents is maintained, where better solutions are used to create new generations of agents using techniques based on biological evolution. Over time, the performance of the population should increase, resulting in strong solutions after many generations of improvement.

Learning-based approaches are currently the most widely used (and arguably most successful) techniques for solving RL problems. The field of deep learning (DL) has impacted RL, rapidly improving the capability of existing RL methods \cite{li2017deep}. Although there has been less interest in evolutionary methods recently, they are starting to return to the limelight, aided by the recent increases in computing power that enabled DL \cite{such2017deep}.

Both learning and evolutionary approaches have been applied to CL, resulting in the development of a range of different techniques for overcoming CF. A recent solution to preventing CF in learning systems is elastic weight consolidation (EWC); it protects the parts of learning systems that are important for the previous problem by adding a penalty for changing them \cite{Kirkpatrick2017}. One evolutionary approach developed agents that regulate their own adaptation rates, meaning they can reduce adaptation in order to preserve performance on the original task \cite{Ellefsen2015}.

With the recent resurgence of research into `deep' evolutionary methods, there is an opportunity to look at the successes of DL and see if they can be applied to evolutionary approaches. This research presents a novel example where a technique (EWC) is successfully taken from a learning approach and applied to an evolutionary approach, with only a small amount of modification. This raises the question, are there other cases where existing learning approaches can be adapted for evolutionary methods?

One may wonder why it is worth investigating evolutionary approaches to RL problems and overcoming CF if learning systems have already been shown to be successful at completing RL and CL tasks. Evolutionary methods have certain advantages over learning systems; these are covered in more detail later in this paper. There is also a large body of evolutionary knowledge that has yet to be exploited with the recent advances in the field, and as the current methods are known to be competitive with DL techniques \cite{such2017deep}, there is potential that evolutionary methods may yet surpass the performance of learning systems.

Additionally, the transfer of information from learning approaches to evolutionary approaches does not have to be in only one direction; there is potential that ideas from evolutionary systems could benefit learning systems. There is also the argument that human intelligence was developed by biological evolution, so mimicking nature's approach to the development of intelligence may not be such a bad idea. In this paper, a novel CL technique is developed to reduce CF when using evolutionary methods. It is based on the EWC technique from learning systems, with adaptation for evolutionary systems.

\section{Background}

When approaching the goal of overcoming CF using evolutionary methods, it is useful to understand the underlying mechanisms that allow individual tasks to be solved independently before considering how to solve several of them sequentially. The review begins with an overview of genetic algorithms (GAs) - one type of evolutionary computing - and explains how they can be used to develop artificial neural networks (ANNs) in a process known as neuroevolution (NE). The second section summarises RL and details some approaches to it - both learning and evolutionary based. Finally, CF is covered, from both an evolutionary and learning perspective, and the main aims of this research are outlined.

\subsection{Neuroevolution}

GAs develop solutions to problems by stochastically simulating the process of natural selection \cite{goldberg1988genetic}. In a GA, a population of individuals is maintained and developed over numerous generations to find increasingly successful solutions. The success of an individual is measured by a fitness function; formally a mapping $f\colon C \rightarrow \rm I\!R^n$ where $C$ is the space of all possible individuals and the fitness of an individual is an $n$-dimensional tuple of real numbers. The fitness of an individual is usually a single real number, but for multi-objective optimisation it is a tuple of size $n$ where $n$ is the number of objectives being optimized for (multi-objective optimisation will be discussed in detail in a later section).

The population of a GA is initially composed of random individuals. Each individual in the population is evaluated and a new population is produced by using techniques based on biological evolution such as crossover and mutation. Crossover is the process of combining two or more existing individuals into a new individual, where fitter individuals are more likely to be selected for crossover. With mutation, a single individual has a chance of changing some of its attributes at random \cite{goldberg1988genetic, mitchell1998introduction}. This basic approach can be applied to any problem that can define a fitness function. For a review of various techniques and applications of GAs, readers are referred to reviews by Davis \cite{davis1991handbook} and Kumar \cite{kumar2010genetic}. 

GAs can be used to evolve ANNs, which are systems of connected artificial neurons (loosely based on neurons in biological brains), where signals are passed between the neurons to produce an overall output \cite{yao1999evolving}. NE is the process of developing ANNs using GAs, where a population of ANNs is maintained and improved over time. Alternatively, gradient-based methods such as back-propagation of errors and stochastic gradient descent allow a network to learn from its mistakes and adjust its connection weights to improve performance on a given task - these are learning based approaches which improve a single network over time \cite{Rumelhart1986}. A generalisation of a feed-forward network is a recurrent neural network (RNN) that includes a feedback loop from previous processing, meaning RNNs have a persistent internal state - a form of memory. In this work, NE is used to develop RNNs.

ANNs with two or more hidden layers are known as deep neural networks (DNNs). Advances in computing power in recent years have made DL (the process of training DNNs) a successful field for developing learning systems \cite{schmidhuber2015deep}. The multiple hidden layers in DNNs allow for complex feature representation and end-to-end learning, with each layer representing increasingly complex abstractions of data \cite{lecun2015deep}. DL has proven very successful when applied to previously unsolved tasks such as image captioning \cite{vinyals2015show} and speech recognition \cite{hinton2012deep}.
 
In a similar way to how gradient-based learning (viz. back-propagation and gradient descent) has benefited from increases in computing power to spawn DL, NE has recently developed into deep neuroevolution (DNE), allowing the  evolution of much larger networks that are competitive with DL techniques \cite{such2017deep}. The recent success for DNE is significant as there are a wealth of techniques that have already been developed for NE that can provide immediate improvement for DNE algorithms, as well as techniques that have been in used DL that may be applicable to DNE.

DNE has some useful advantages over DL; primarily that it can evolve the ANN architecture at the same time that it evolves the weights of the connections. In conventional DL, the ANN architecture is chosen by a human designer who bases the structure on past experience and domain knowledge. This often means multiple networks must be developed with different architectures to determine the optimal layout. However, with topology and weight evolving artificial neural network (TWEANN) algorithms, GAs are used to find optimal network architecture.

\subsection{Reinforcement learning}

RL is the process of learning via interaction within an environment, where a system develops a strategy to maximise some reward by gaining experience through trial and error \cite{sutton1998reinforcement}. RL can be applied to physical and virtual environments, with robots acting in the physical world for the former and software agents acting in simulations for the later. Progress in deep learning has benefited RL, spawning the new field of deep reinforcement learning (DRL) \cite{li2017deep}. This has furthered success on much more difficult problems in higher dimensional spaces, such as superhuman performance on Atari 2600 video games \cite{mnih2015human} and Deepmind's accomplishments in the domains of Go and chess \cite{silver2016mastering, silver2017mastering, silver2017mastering2}.

\begin{figure}[h]
	\includegraphics[scale=0.9]{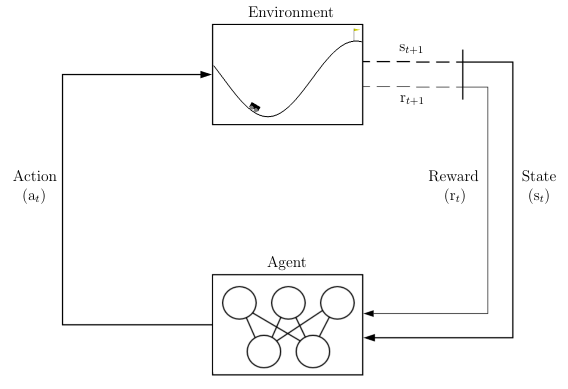}
	\centering
	\caption{The perception-action loop. The agent observes a state s$_t$ from the environment at time t and performs an action a$_t$. The environment transitions into the next state s$_{t+1}$ and gives the agent a reward r$_{t+1}$ for the action it just performed.}
	\label{fig:perception-action_loop}
\end{figure}

The aim of an RL agent is to maximise a reward for a task, such as the overall score in a video game. An agent can increase its reward by taking actions within its environment, which will move the environment and the agent into a new state based on the chosen action and the current state. The environment then rewards the agent based on the action it just took. An agent decides on which action to take by observing its current state (this can be fully or partially observable) and by following a control policy which is developed over time by the learning system. This can be represented as a perception-action loop (see Figure \ref{fig:perception-action_loop}).

RL can be formulated as a Markov decision process; this allows formal definitions of the control policies that RL systems create. A Markov decision process has five components:

\begin{itemize}
	\item A set of states $S$.
	\item A set of actions $A$.
	\item A transition function $\mathcal{T}(s_{t+1}|s_t,a_t)$ that describes the state reached at time $t+1$ if action $a$ is performed when in state $s$ at time $t$. This can be adapted such that a state-action pair maps to a probability distribution over a set of states - this is useful in environments with uncertainty.
	\item A reward function $R(s,a,s')$ for transitions from state $s$ to state $s'$ by performing action $a$.
	\item A reward discount factor $\gamma \in [0,1]$ that defines the priority of immediate reward against long term reward - a lower values places emphasis on immediate reward. 
\end{itemize}

The control policy $\pi$ which the agent follows is simply a function from states to actions, formally $\pi \colon S \rightarrow p(A = a|S)$. The mapping is to a probability distribution over actions as, in some cases, the agent may want to make actions stochastically. The aim of the learning system is to find an optimal policy $\pi^*$ that maximises expected reward over time:
\begin{equation}
	\pi^* = \underset{\pi}{\operatorname{argmax}}\, \mathbb{E}[R|\pi]
\end{equation}

An extension of the Markov decision problem is the partially observable Markov decision problem (POMDP), in which an agent only receives a limited amount of information about the environment \cite{kaelbling1998planning}. POMDPs are very common in physical environments in which the agent only has information from its sensors and not from the environment itself. A common technique for dealing with POMDPs is to use RNNs as they retain information observed from previous states and can therefore develop a better understanding of the environment \cite{hausknecht2015deep}.

There are two main techniques for approaching RL problems: value function search and policy search. The former is based on estimating the value (expected overall reward) of being in a given state; this allows optimal action selection by evaluating the states that can be reached from the current state and choosing the next state that has the greatest value. The values of each state are not initially known, therefore the system must improve its understanding of the task by developing an internal model of state values. This is the basic concept behind Q-learning and by extension deep Q-learning \cite{mnih2015human, watkins1992q}. Policy search, on the other hand, is a model-free method - it has no internal model of state values. Rather than create a state value function model, it searches the space of policies. ANNs can encode policies, which allows training by either gradient-based or gradient-free methods \cite{arulkumaran2017brief}. NE can be used as a gradient-free policy search method; the fitness function of the GA is simply the reward function of each individual's performance on the RL task.

\subsection{Catastrophic forgetting}

The ability to learn new skills on top of existing ones is a crucial concept for the development of artificial general intelligence. Due to the complexity of the real world, it is impossible to train for all possible scenarios, therefore systems must be adaptable and able to perform learning beyond their initial training. This is the essence of CL - a system that is able to perform one task must learn a novel task while retaining its ability in the original task; unfortunately, conventional methods for developing ANNs are very poor at doing this \cite{MCCLOSKEY1989109}.

CF is the act of overwriting what has been previously learnt - the system becomes competent at a new task but forgets some or all of its knowledge about the previous task. When ANNs are trained on a new task, they adjust their connection weights, but as all the weights were finely tuned to solve the original problem, a major adjustment to the weights in order to solve the new problem can dramatically reduce performance on the original task \cite{ratcliff1990connectionist}.

Several solutions have been proposed to negate CF in ANNs, for example activation sharpening \cite{french1992} and orthogonalization \cite{French1994} (for a review of more methods see \cite{French1999}). In nature, CF is a very rare occurrence, implying biological evolution managed to find a solution to this problem \cite{French1999}. The aforementioned solutions were all engineered, therefore are not representative of how CF was avoided in biological evolution, hence the proposal for an evolved solution by Ellefsen et al. \cite{Ellefsen2015}.

An investigation into why biological networks are resistant to CF finds a dependence on modularity \cite{Hintze2008}. Modular networks have regions with high connectivity that are sparsely connected to other regions. In order to try negating CF by using modularity, modular networks must first be evolved. One technique for evolving modularity is to expose the system to changing environments over time; an effective example is using modularly varying goals (swapping between multiple different goals that are composed of common sub-tasks over the course of an evolutionary run) \cite{Kashtan2005, Kashtan2007}. This is also successful in speeding up the rate of evolution. However, there are some disadvantages to this method, notably that it is not effective with randomly varying goals (goals with no common sub-tasks) and that is requires repeated exposure to the original task in order to remain competent in it. Therefore, the use of modularly varying goals is not a scalable and realistic solution to catastrophic forgetting \cite{Ellefsen2015}. 

Another flaw with modularly varying goals is that they are not that representative of the environment for natural evolution: the rate of goal swapping required to produce modularity is far greater than is likely to occur in nature and decomposition of goals in nature is not always entirely obvious \cite{Ellefsen2015}. Alternative research by Clune et al. \cite{Clune2013} found that applying selective pressures towards networks with a reduced number of connections causes modularity in both changing and unchanging environments, thus removing the need for interleaving the training tasks and continuous revisiting of older tasks. 

On its own, modularity is only partially effective for overcoming CF; another mechanism is required to `protect' neurons that are vital for a previously learnt task. In biological neural systems, chemicals are released in the brain that control the learning rate of neurons in the system - either making them quicker or slower to change their connection weights. It is possible to replicate this in ANNs by using modulatory neurons that up or down regulate the learning rates of neurons they are connected to \cite{Soltoggio2008}. This can be applied to CF by using modulatory neurons to protect neurons that are used in one task by down regulating their learning rates when learning a new task \cite{Ellefsen2015}. Unfortunately, this method uses a style of network that is different to conventional ANNs due to the inclusion of special modulatory neurons, so cannot be easily integrated with existing NE techniques. Also, since some of the neurons in the networks are modulatory neurons that do not contribute to processing, only regulation of learning rates, the networks can end up being larger than conventional ANNs when applied to the same task. Therefore there is an incentive to look for an alternative evolutionary method for reducing CF that does not require the use of modulatory neurons. 

The techniques mentioned above are only focused on preventing CF when using evolutionary methods. A recent study by Kirkpatrick et al. \cite{Kirkpatrick2017} produced the EWC technique for overcoming CF when using DL (as opposed to GAs/NE). The technique adds a penalty to the loss function of the learning system for changing the weights of connections that are important to the original task. This proved very successful, showing significant prevention of CF when applied to DRL, allowing a single network to master several Atari 2600 games, albeit at a lower performance level than what was achievable when only training on a single game. 

Thus far, no attempt has been made to apply EWC to evolutionary methods. Although some aspects of the technique rely on information that is readily available when training via learning (but not when training via evolution), the high-level idea of directly penalising weight changes that affect connections that are used in the original task when training on a new task is transferable to evolutionary methods. The aim of this research was to develop a novel technique that is based on EWC but used with NE rather than learning systems.

\section{Experimental Setup}

This section covers the experimental techniques and implementations used in this work. It begins by defining the novel Weight Protection technique for reducing CF when using evolutionary methods. The evolutionary strategy is then outlined, followed by a description of the RL tasks that were used. Finally, the implementation details are outlined.

\subsection{Weight protection}

In order to transfer the EWC technique developed by Kirkpatrick et al. \cite{Kirkpatrick2017} to evolutionary methods, further analysis of the method must be undertaken. The idea of the technique is to penalise changes to the connection weights of the network after learning the first task. Certain connections will be more important to the original task than others, therefore the penalty for changing them should be greater. As a result, the network should remain similar to its original configuration once the second task has been learnt, meaning it should retain its performance on the first task. This method exploits the fact that there are a range of configurations that will give a high level of performance on a task, and there is likely to be an overlap in the solution spaces of the tasks to be learnt.

EWC was originally created for use with DL, therefore it is implemented via the loss function. The loss function can be defined as a function $L_T(\theta)$, where $T$ indicates a task specific loss function and $\theta$ represents the model parameters, i.e. the network connection weights and neuron biases. The revised loss function for EWC is:

\begin{equation} \label{eq:ewc}
L(\theta) = L_B(\theta) + \sum_i \frac{\lambda}{2} F_i (\theta_i - \theta^*_{A,i})^2
\end{equation}

where $i$ is used for iteration over each of the model parameters, $\theta^*_{A,i}$ are the model parameters after learning task A, $F$ is the Fisher information matrix and $\lambda$ is a parameter to define the importance of task A against task B. The summation in the equation is the penalty term introduced by EWC. The Fisher information matrix represents the importance of each of the model parameters, i.e. which network connections are most crucial for high performance on task A. This means changing less important connections will not increase the penalty term as much as changing more important connections.

When considering how to apply the EWC formula to NE, there are two hurdles to overcome: 
\begin{itemize}
	\item Fisher information matrix - This can be calculated using the first derivatives of the task specific loss function with respect to each parameter. These first derivatives are required when using gradient-based learning techniques as they dictate the direction in which each model parameter is adjusted. In certain cases, the first derivative calculation is intractable, so approximations must be made. However, with gradient-free techniques (e.g. NE), the first derivative is not used, which is advantageous for functions with intractable derivatives. Therefore, when transferring the EWC technique to NE, it would be beneficial if the method worked without having to calculate the first derivatives, meaning the Fisher information matrix cannot be used.
	\item Model comparison - When using learning based methods, a single network is developed over time, whereas evolutionary methods maintain a population of networks. The EWC method requires comparison to an optimal network for the original task, and with learning methods this is an easy process since there is only a single network to compare to. With evolutionary techniques, there are multiple networks that could be chosen (e.g. the smallest network, the fittest network, the network that changes the weights the least etc.), so the various options must be considered when transferring the EWC method to evolutionary techniques.  
\end{itemize}

Weight Protection (WP), the novel technique designed for this research, directly tackles both of these issues. The solution to the first problem (removing the use of Fisher information yet still prioritising the protection of certain weights) utilises the ability of NE to `grow' networks by applying one penalty to existing connections and another penalty to new connections. This reasoning assumes that all of the existing connections are equally important to the original task and that new weights are of a different importance. This gives a new loss function (or rather a fitness function since it is now used for evolutionary methods):

\begin{equation} \label{eq:wp}
F(\theta) = F_B(\theta) - \sum_i \lambda_1 (\theta_i - \theta^*_{A,i})^2 - \sum_j \lambda_2 (\theta_j)^2
\end{equation}

where $i$ is used to iterate over existing connections, $j$ is used to iterate over new connections, and $\lambda_1$ \& $\lambda_2$ define the relative importance of each penalty term. Note that in Equation \ref{eq:wp} the signs for the penalty terms have swapped from Equation \ref{eq:ewc}; this is because evolutionary methods normally try to maximise a fitness function rather than minimise a loss function.

Both Equation \ref{eq:ewc} and \ref{eq:wp} have quadratic penalty terms; for EWC this is described as ``a spring anchoring the parameters to the previous solution''\cite{Kirkpatrick2017}. The quadratic parameters are comparable to $l_2$ regularisation, a technique usually used to prevent over-fitting of networks by penalising connections with high weights, aiming to reduce to the complexity of the model. In effect, the second penalty term of the revised fitness function for WP is just $l_2$ regularisation of the newly added connections.

WP uses the individual that has the highest revised fitness as the comparison model. There is a trade-off between getting a high level of fitness on the new task and keeping the model as similar to the original model as possible. The revised fitness function of WP incorporates both of these aspects, so an individual that has a high score on the revised function should provide a good balance of both requirements.

In addition to using a revised fitness function, WP can be combined with methods designed to induce modularity in networks, which improves task retention \cite{Ellefsen2015}. Multi-objective optimisation allows for simultaneous improvement of two or more objectives. In the case of WP, the objectives are to maximise the revised fitness function and to minimise the network sizes (number of connections). This follows a similar technique used by Ellefsen et al. \cite{Ellefsen2015} to create modular networks, where the second objective of minimising network size is less important, therefore is only applied with a probability $p$, where $0 \leqslant p \leqslant 1$. If $p = 1$ then the second objective is as important as the first, if $p = 0$ then it is entirely ignored. 

When optimising for several objectives, it is likely to be impossible to maximise all of them - there is a trade-off in performance across the different tasks. Consider two solutions to a multi-objective optimisation problem, $s_1$ and $s_2$, where $s_1$ has a better performance across all objectives than $s_2$. Clearly, $s_1$ is a better solution than $s_2$; $s_1$ is said to dominate $s_2$. However, if $s_2$ is better than $s_1$ in just one of the objectives, then they are equally optimal solutions. A Pareto-optimal solution is a non-dominated solution: one that cannot be changed such that performance increases for all objectives. The set of all Pareto optimal solutions is known as the Pareto front, which represents the limit of optimality across all objectives \cite{steuer1986multiple}. 

During the development of the WP method, an alternative implementation of the WP penalty was tested. Rather than include the penalty term in the fitness function, it became another objective to optimise and the overall fitness function was just the fitness function for the current task. However, this produced a three-dimensional Pareto front, and the majority of the individuals in the population did not have the desired features, so this method was discarded. It was found that including the penalty term in the fitness function gave much better results. 

With the revised fitness function and optimisation for modularity, WP has three tunable hyper-parameters: $\lambda_1$, $\lambda_2$ and $p$. This seems to be more hyper-parameters than EWC, but since evolutionary methods can change network topologies as well as connection weights, there are fewer things to configure when using WP. Under normal circumstances, $\lambda_1$ should be greater than $\lambda_2$ as, in order to preserve the performance on the original task, the penalty for changing existing weights should be greater than the penalty for creating new connections.

\subsection{Evolutionary strategy}

When using ANNs to define policies in RL problems, RNNs are more useful than feed-forward networks as they have a form of memory from previous processing and can be successfully applied to POMDPs, which feed-forward networks cannot. Therefore, when looking at generic networks for solving RL problems, it is better to use RNNs. In this research, the novel WP technique was used to evolve RNNs because of their wider application area - successful utilisation of WP for RNNs implies greater potential across a wider range of domains. The RNNs used in this research were fully connected RNNs - the neurons in the hidden layer(s) and output layer had recurrent connections to every neuron within their layer. The RNNs used the $tanh$ activation function at every level, which has an output of $(-1,1)$ meaning the recurrent signals do not explode over time.

The evolutionary strategy used a TWEANN GA with a multi-objective optimisation selection method. The TWEANN algorithm was based on the neuroevolution of augmenting topologies (NEAT) method \cite{stanley2002evolving} (the most widely used TWEANN algorithm) but was more lightweight and was adapted to make controlling the size of networks easier. Networks began with very few connections and grew over time, but the network architecture had a fixed pattern to restrict the maximum network size, i.e. there was no addition or removal of nodes, only weights. To encourage varying network sizes, a mutation could add or remove random weights as well as adjusting the connection weights, as per the NEAT method. The chosen multi-objective optimisation selection method was the non-dominated sorting genetic algorithm (NSGA-II) \cite{Deb2002} - a popular canonical method for finding Pareto-optimal solutions when using GAs. NSGA-II selects individuals from the population based on maximum diversity across the Pareto front. 

Individuals were encoded as fixed sized arrays of floating point numbers. NEAT usually has to deal with different sized individuals, but since the networks had a fixed architecture (with lots of zero weights initially so that networks started off small), the algorithm used in this experiment did not require the capacity for performing crossover on different network architectures. The encoding was a direct genotype to phenotype mapping, with each floating point number representing a single parameter of the network. 

Since the focus of the research was to test the WP method, certain evolutionary hyper-parameters were kept the same across all evolutionary runs to ensure consistency. The experiments used single point crossover: a single random point is chosen somewhere along the genotype of both parents, and all information beyond that point is swapped between the two parents \cite{mitchell1998introduction}. The probability of crossover was set at $0.6$. Gaussian mutation was also used - if an individual is selected for mutation then each gene in the individual has a one in ten chance of being mutated. The mutation of a single gene was performed by drawing a value from a Gaussian distribution (with a mean of zero and a standard deviation of one) and adding it to the existing value of the gene \cite{hinterding1995gaussian}. The probability of an individual being selected for mutation was 0.4. As previously mentioned, an additional mutation strategy was included to add or remove a random connection. The probability of adding or removing a connection when mutating an individual was 0.1, with a uniform probability across all connections. The population had a fixed size of 100 individuals, and used the $(\mu + \lambda)$ evolutionary technique to produce 100 offspring and then select the 100 best individuals from the population and the offspring to form the next generation. 

In addition to keeping the crossover and mutation conditions consistent across all tasks, the modularity optimisation parameter $p$ was also kept the same. It was implemented by only applying the second optimisation objective every $1/p$ generations. Initial tests were used to find the value of $p$ that was most effective at reducing CF while just using modularity optimisation, i.e. without using the WP method.

\subsection{Reinforcement learning tasks}

Several different RL tasks were used to test the WP technique. This made use of the OpenAI gym: a library that provides a collection of different RL environments, from simple text-based tasks to Atari 2600 games \cite{brockman2016openai}. In order to facilitate comparison between the tasks that were used, a minimum and maximum fitness was found for each task by performing several single task evolutionary runs. This meant a normalised fitness function in the range $[0,1]$ could be calculated for each task. 

The tasks had random starting conditions, meaning each run of a task would begin slightly differently. To ensure consistency when running the tasks and to make sure the developed agents were competent from various starting positions, five different seeded starting positions were used for each task. This meant that one `run' of a task actually meant running it five times from different starting positions. 

Each of the RL environments used in this work are classic control tasks - benchmarks that have previously been used in the literature for comparing different techniques. Since the progression to DRL, these tasks have become less useful benchmarks because they can be solved with relative ease - modern benchmarks have progressed onto harder tasks such as the Atari 2600 games. However, for the scope of this research, they still provide a suitable standard for testing CL. Along with four tasks from the OpenAI gym library (cart-pole balancing, pendulum swing-up, acrobot and mountain car), an additional task was developed that was specifically design as a POMDP to test the `memory' of the evolved RNNs. The task was based on simple experiments involving orientation and discrimination of shapes \cite{Beer1996}. In these experiments, the agent has an array of five distance sensors and can only move horizontally. For the chosen task, the agent has to track falling circles, however, the circles move at a greater horizontal velocity than the agent, therefore the circles will pass beyond the agent's sensors. The agent can catch up when the circle collides with the edge of the environment, but it has to remember the direction the circle was travelling in, hence the requirement for an RNN. 

\subsection{Implementation}
\label{sec:implementation}

All of the code for this research was written in Python. While there are Python machine learning libraries that have RNN implementations, they cannot be easily integrated with evolutionary methods, therefore a custom RNN implementation was written from scratch. The GAs were implemented using DEAP, a Python library for evolutionary algorithms \cite{DEAP_JMLR2012}. As mentioned above, the OpenAI gym library provided the RL tasks, however, it did not include the Cognitive Agent Orientation task. This was implemented as an extension to the library, which exposes the necessary features that allow the addition of custom tasks.  

The OpenAI gym uses two types of output spaces - discrete and box. For discrete outputs, one signal out of $n$ total signals is set to one, with the rest set to zero (one-hot encoding). For box outputs, each signal is a continuous value somewhere between a minimum and maximum value; different signals can have different minimum and maximum values. Box signals are also used as inputs to the tasks. For a description of the input and output spaces for each of the RL tasks used in this work, see Table \ref{tab:evns}. 

\renewcommand{\arraystretch}{1.2}
\begin{table*}
	\centering
	\begin{tabular}{C{3cm}|C{2.2cm}|C{2cm}|C{6cm}}
		\textbf{Task} & \textbf{Input Size} & \textbf{Output} & \textbf{Output description} \\
		\hline 
		Cognitive Agent Orientation & 5 & Discrete 3 & Move left, do nothing, move right  \\
		\hline
		Cart-Pole Balancing & 4 & Discrete 2 & Move left, move right \\
		\hline 
		Pendulum Swing-up & 3 & Box 1 & Amount of torque to apply; sign determines direction \\
		\hline 
		Acrobot & 6 & Discrete 3 & Apply counter-clockwise torque, do nothing, apply clockwise torque \\ 
		\hline 
		Mountain Car & 2 & Discrete 3 & Reverse, do nothing, accelerate \\
	\end{tabular} 
	\caption{Outputs and descriptions for the RL tasks used in this work.}
	\label{tab:evns}
\end{table*}

Using an ANN to make actions in RL tasks requires interpretation of the network's output for the task it is currently applied to. In this work, the network outputs are always in the range $(-1, 1)$ since the output neurons use the $tanh$ function. For discrete outputs, the index of the maximum output of the network is the action taken, i.e. a network output of $\{-0.5, 0.8, 0.3\}$ for a discrete three signal output would result in the second action being taken. With a box output, each of the relevant signals is scaled to the range defined by the task, i.e. for a box single signal output with minimum and maximum values $\{-2,2\}$, a network output of $-0.5$ would be scaled to $-1$. 

When applying one network to multiple tasks, the tasks are likely to have different sized output spaces, therefore some output signals are ignored for certain tasks. In this work, the network outputs were kept from the left, i.e. if a network with an output size of four was applied to a three output problem, the right-most output was discarded. For this work, networks had to have an input layer of size six and an output layer of size three in order to be able to do all the tasks, with an arbitrary configuration of hidden layers.

\section{Results}

\subsection{Modularity with continual task swapping}

Before using the WP method, the best modularity parameter $p$ was found. Evolutionary runs were performed with changing tasks, where once the average fitness of the population had reached $95\%$ performance on a task, the focus shifted onto the next task, repeating in a loop until the population had a performance of $95\%$ on all tasks at once. Three tasks were used: cart pole, mountain car, and agent orientation. The best value of $p$ that was found was $0.2$, i.e. applying the network size penalty every five generations. This resulted in the fastest evolution (least number of generations) and typically each task was revisited five times within an evolutionary run. While this method was successful in developing a network that could do all three tasks, it would not be effective on a larger scale with more tasks as each task must be revisited several times. 

\subsection{Establishing weight protection}

\begin{figure}
	\includegraphics[scale=0.38]{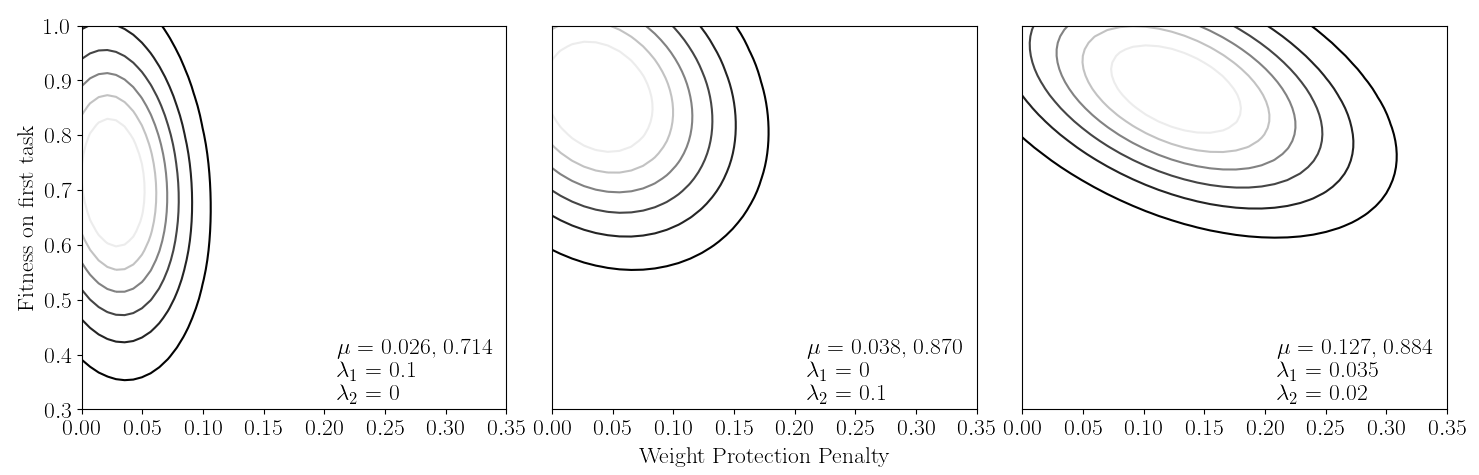}
	\centering
	\caption{A comparison of the correlation between the WP penalty and the fitness on the first task after learning the second task for three different configurations of $\lambda_1$ and $\lambda_2$. Each Gaussian plot was produced from 30 evolutionary runs, where the final population of each run was re-evaluated on the first task and the WP penalty. }
	\label{fig:gaussian_comp}
\end{figure}

To establish the ability of WP to reduce CF, a useful starting point was to check there was a negative correlation between the WP penalty (the two summation terms in Equation \ref{eq:wp}) and the fitness on the original task after learning the second task, i.e. a lower penalty implies less forgetting, so minimising the penalty is beneficial for reducing forgetting. Several repeats were completed with different configurations of $\lambda_1$ and $\lambda_2$ (the two hyper-parameters for WP) to verify the WP method and to also find the best hyper-parameter settings for later tests. The modularity pressure was used in all of the evolutionary runs, with $p$ set to 0.2 as described in the previous section. The evolutionary process consisted of evolving the population on the pendulum swing-up task for 40 generations, and then on the acrobot task for another 40 generations; the original task was never revisited once it had been learnt. As demonstrated by Figure \ref{fig:gaussian_comp}, use of both $\lambda_1$ and $\lambda_2$ gives the strongest negative correlation, implying the WP method is effective at preventing forgetting. This configuration also has the highest average fitness on the first task after learning the second task, which further supports the use of both WP parameters. 

\begin{figure}[h]
	\includegraphics[scale=0.7]{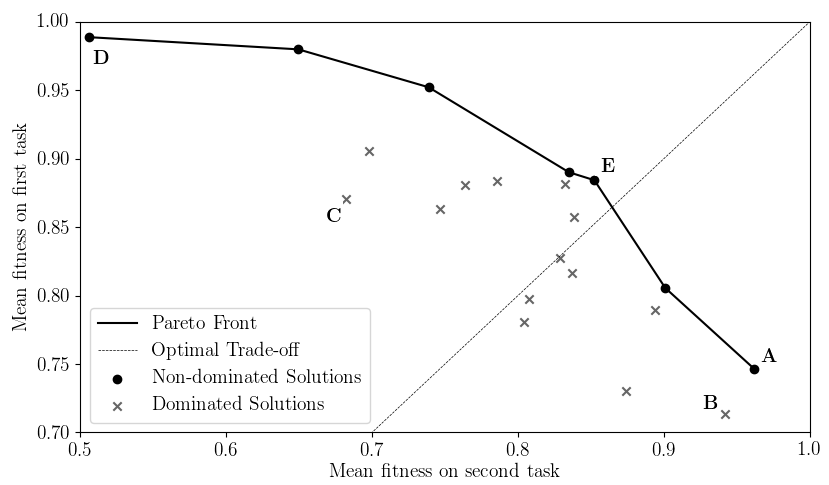}
	\centering
	\caption{The mean performance on task one and task two for different configurations of the WP hyper-parameters averaged across 30 different runs. The Pareto front shows the best solutions found, and the optimal trade-off line shows where equal performance on both tasks would be. Some important data points are highlighted:
	A) No WP B) $\lambda_1$ only C) $\lambda_2$ only D) $\lambda_1=0.2, \lambda_2=0.1$ - penalty terms too large. Excellent retention of the original task but poor performance on the second task.
	E) $\lambda_1=0.035, \lambda_2=0.02$ - the best configuration found. Strong performance on both tasks and quite close to the optimal trade-off line. }
	\label{fig:trade_off}
\end{figure}

Adjusting $\lambda_1$ and $\lambda_2$ varies the task priority between preserving performance on previously seen tasks and gaining performance on new tasks. The focus when choosing the WP hyper-parameters in this research was on an optimal trade-off between the two tasks, i.e. maximising performance on both. A larger penalty places priority on retaining the performance on the old task, a smaller penalty increases forgetting but increases performance on the new task. The balance between the two parameters can be tweaked to fine tune the trade-off - configurations that produced a strong mean performance on both tasks were found to have $\lambda_1$ between 1.5 and 2.5 times greater than $\lambda_2$. This verifies the earlier claim that $\lambda_1$ should be greater than $\lambda_2$ to encourage networks to add new connections rather than change existing ones. Figure \ref{fig:trade_off} shows the performance of various configurations of $\lambda_1$ and $\lambda_2$.

\subsection{Evaluating weight protection}

The aim of the WP technique is to improve retention of previously learnt tasks when learning new ones. The success of the novel WP technique is analysed in two different ways below, namely final performance on all the tasks after learning and drop-off in performance on tasks as subsequent tasks are learnt. For both analysis techniques, a comparison was done between four different evolutionary methods: normal evolution, evolution with modularity, evolution with WP, and evolution with WP \& modularity. The hyper-parameters used were kept consistent across all experiments: $p = 0.2$, $\lambda_1 = 0.035$ and $\lambda_2 = 0.02$; these values are the best configuration found from the previous experiments. An additional parameter in the experiment was the maximum network size. In this instance, it was limited to a single hidden layer with up to six hidden nodes, resulting in 108 model parameters to be tuned via evolution.

\begin{figure}[h]
	\includegraphics[scale=0.8]{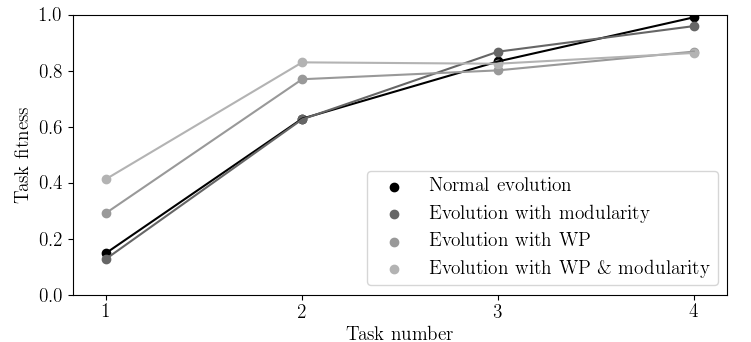}
	\centering
	\caption{The mean performance of the best individuals from 30 evolutionary runs of different evolutionary techniques applied to four different tasks. Evolution without WP results in CF of previously seen tasks, yet better performance on the most recently seen task (hence better performance on task four for non-WP methods). }
	\label{fig:task_performances}
\end{figure}

When applied to four tasks (cart pole, pendulum swing-up, mountain car and agent orientation), WP showed a significant improvement in the overall performance across all four tasks, both with and without using modularity (see Figure \ref{fig:task_performances}). As expected, it had lower performance in the tasks than could be achieved by solely evolving for each one individually, but this is the trade-off for retaining performance on the other tasks. As per the previous experiment, the population was evolved on each task for 40 generations before moving onto the next task, resulting in 160 generations of evolution in total. After evolution, the individual with the highest fitness on the final task was re-evaluated on all four tasks. Note that the best individual was chosen based on its final fitness, not by re-evaluating the entire population on all tasks and choosing the one with the best performance across all of them. This meant the method of selection of the best individual became dependent on the evolutionary method currently being used: normal evolution would choose the individual with the highest final task fitness, yet evolution with WP would choose the individual with the highest revised fitness, i.e. not necessarily the individual with the best performance on the final task, but the best performance across all tasks.

Since the evolutionary methods are stochastic, they produce different results each time. This means there is inherent variability in the results, hence why many evolutionary runs are performed for each experiment. Due to this variability, it is useful to compare the best results for each technique as well as the mean results (see Table \ref{tab:task_perf}). In this experiment, the best run was considered to be the run that produced the highest average performance across all tasks. While WP does provide a noticeable improvement to the average overall performance ($\sim$0.083), the improvement for best overall performance is more significant ($\sim$0.143). 

\renewcommand{\arraystretch}{1.2}
\begin{table}[h]
	\centering
	\begin{tabular}{C{2.5cm}|C{2cm}|C{2cm}|C{2cm}|C{2cm}|C{2cm}}
 \textbf{Technique} & \textbf{Task 1} & \textbf{Task 2} & \textbf{Task 3} & \textbf{Task 4} &\textbf{Overall}\\ \hline
Normal &  0.151/1.000  &  0.630/0.472  &  0.834/0.798  &  0.992/1.000  &  0.651/0.818 \\ \hline
Modularity &  0.130/0.360  &  0.627/0.939  &  0.869/0.851  &  0.960/1.000  &  0.647/0.787 \\ \hline
WP &  0.294/1.000  &  0.771/0.896  &  0.802/0.941  &  0.869/1.000  &  0.684/0.959 \\ \hline
WP \& Mod. &  0.415/1.000  &  0.831/0.957  &  0.826/0.917  &  0.864/0.971  &  0.734/0.961
	\end{tabular} 
	\caption{Performance on each task for the four different evolutionary methods. Each cell contains the task fitness averaged across the 30 runs followed by the task fitness from the best run.}
	\label{tab:task_perf}
\end{table}

\begin{figure}[h]
	\includegraphics[scale=0.8]{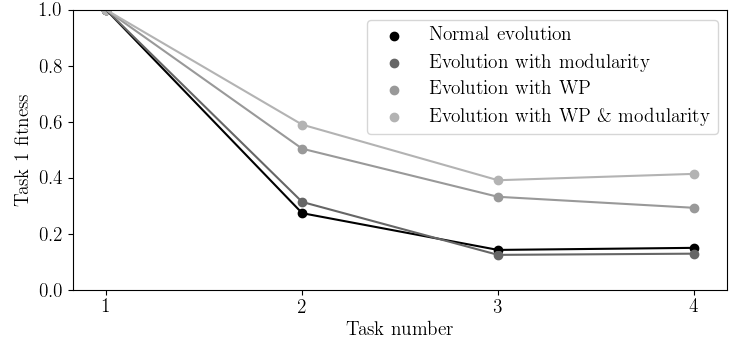}
	\centering
	\caption{The drop-off in performance on task one as each new task is learnt, averaged over 30 evolutionary runs per technique.}
	\label{fig:task_dropoff}
\end{figure}

It is also insightful to look at the drop-off in performance on a task as other tasks are learnt. In theory, as more tasks are learnt, the original task performance will decrease, and better methods will reduce the rate at which the performance decreases, i.e. slow the rate of forgetting. By analysing the performance on the original task at every task swap in the previous evolutionary experiment, the drop-off rate can be compared for each of the different evolutionary techniques (see Figure \ref{fig:task_dropoff}). It is clear that the use of WP significantly reduces the decrease in performance, implying it is successful at reducing forgetting. The inclusion of modularity also helps to alleviate forgetting, but only when using in conjunction with WP - on its own it provides no discernible improvement.

\section{Conclusion}

Overall, the research undertaken in this project has successfully demonstrated a novel technique for reducing CF when evolving ANNs using GAs. It has been shown, through various different experiments and types of analysis, that the technique improves retention of previously learnt tasks when compared to normal evolution. 

Due to time and resource constraints, the novel technique was only tested against relatively easy and slightly out-dated RL tasks. To better evaluate the potential of the novel WP technique, it would need to be applied to more complex tasks. In addition, the WP technique was not implemented with the full NEAT method (a more lightweight version was created instead), so there is still room for further development and testing of WP with other methods.

It was beyond the scope of this research to compare the technique to existing methods for overcoming CF, both learning and evolutionary based. While the novel technique certainly shows improvement when compared to normal evolution, its overall usefulness could be further evaluated with direct comparison to existing techniques such as EWC. 

On a broader spectrum, the development of the WP technique has demonstrated a successful transfer of knowledge and an approach (EWC) from an application in learning systems to an application in evolutionary systems, with only a relatively small amount of adaptation. This could highlight that there is a further possibility for techniques from both domains to be shared with one another, and potential development of hybrid techniques using the best of both areas of research. 

An additional area of further research could be an extension of the WP method that revisits tasks. Qualitative analysis of the tasks that were used in this research found that some were more prone to CF that others. An interesting hybrid technique could learn all the tasks using the WP method, then revisit only the tasks that showed a significant decrease in performance. This method would potentially increase the overall task fitness, yet retain the scalability of WP.

\bibliographystyle{ieeetr}
\bibliography{bib}

\end{document}